# Reliability-Guided RGB-D Sensor Fusion for Glare-Resilient Navigation Costmaps

**Shang-En Tsai**

***Abstract*—**Specular glare on reflective floors, glass boundaries, and glossy indoor surfaces can corrupt active-stereo red-green-blue-depth (RGB-D) measurements, producing holes and spikes that persist as phantom obstacles in navigation costmaps. This paper presents a glare-resilient RGB-D sensor-fusion method based on explicit per-pixel depth reliability. A lightweight Depth Reliability Map network (DRM-Net) predicts sensor trustworthiness, and reliability-guided fusion (RGF) combines continuous weighting with a minimum gate before occupancy integration. Rejected measurements generate neither obstacle insertion nor free-space clearing; affected cells remain unknown or retain prior evidence. Training targets are built from a five-frame, pose-aligned multi-view buffer using independent LiDAR/AMCL poses, occlusion-aware aggregation, and a calibrated range-dependent depth-uncertainty model. The evaluation includes tuned nvblox TSDF, Intel RealSense SDK post-processing, high-threshold stereo-confidence filtering, TDCNet, and HDCNet baselines, together with statistical, safety, generalization, and embedded-runtime analyses. Under severe glare, DRM+RGF achieves FOR 0.056±0.012, FSR 0.897±0.045, FNOR 0.018, 1.00±0.00 degraded-mode safety interventions per 10 m, and 91.4% task success, while operating at 16.5 ms per frame. Across the retained reflective-scene trials (Baseline N=98; DRM+RGF N=105), collisions decrease from 14 to 1. These results support reliability-guided fusion as a favorable safety–utility tradeoff relative to aggressive filtering and dense completion for glare-affected indoor navigation.

## I. Introduction

Occupancy-grid costmaps remain a core representation for mobile robot navigation because they are simple, interpretable, and compatible with real-time planners. Their reliability, however, depends critically on upstream depth measurements. In reflective indoor environments, such as polished floors, glass partitions, glossy tiles, and wet or coated surfaces, specular glare can corrupt RGB-D depth sensing by introducing invalid measurements, depth holes, and large positive spikes. Once integrated over time, these corrupted measurements may become persistent phantom obstacles, causing unnecessary detours, repeated recovery behaviors, emergency stops, and occasionally unsafe motion decisions.

Recent work addresses reflection-induced RGB-D failures through reflection-aware mapping, reflective-ground SLAM, sensor calibration, depth-error modeling, learned depth completion, and uncertainty-aware fusion [1]–[18]. These directions are complementary, but they optimize different objectives. Completion methods seek dense geometry; mapping filters reject inconsistent measurements; and uncertainty-aware methods attenuate evidence according to estimated confidence. For local navigation, however, a visually plausible reconstruction is not sufficient: the system must limit both phantom-obstacle insertion and false free-space creation.

In this work, glare resilience is treated as an RGB-D depth-reliability problem. The Intel RealSense D435 projects an infrared pattern and estimates depth through stereo correspondence [7]. On polished floors and glass-like surfaces, specular and multipath reflections can distort this pattern, causing depth holes and large disparity spikes. These failures motivate per-pixel reliability estimation before map integration rather than post-hoc map repair.

The proposed RGB-D sensor-fusion system uses DRM-Net to estimate a Depth Reliability Map (DRM). The default five-channel input comprises RGB, measured depth $D_t$, and the temporal depth difference $|D_t - D_{t-1}|$. The resulting reliability map is used by Reliability-Guided Fusion (RGF), which combines continuous reliability weighting with a minimum reliability gate. In all experiments, severely corrupted depth-sensor measurements are blocked from obstacle insertion, while valid measurements above the threshold retain graded reliability and contribute proportionally to occupancy updates. The complete pipeline is summarized in Fig. 1.

The evaluation is organized around five questions: (1) whether pose-aligned, independently localized multi-view references reduce circular supervision; (2) whether the learned reliability target outperforms binary confidence and classical filtering; (3) whether reliability-guided fusion offers a better FOR–FSR–FNOR tradeoff than RealSense post-processing, stereo-confidence filtering, and representative depth-completion networks; (4) whether the gains remain statistically significant and generalize to an unseen reflective material; and (5) whether the method reduces collisions and near-collisions while maintaining embedded real-time throughput. In summary, estimating depth reliability before occupancy integration suppresses glare-induced phantom obstacles without replacing uncertain measurements with invented geometry.

Shang-En Tsai is with the Department of Computer Science and Information Engineering, Chang Jung Christian University, Tainan 711301, Taiwan (e-mail: sean@mail.cjcu.edu.tw).

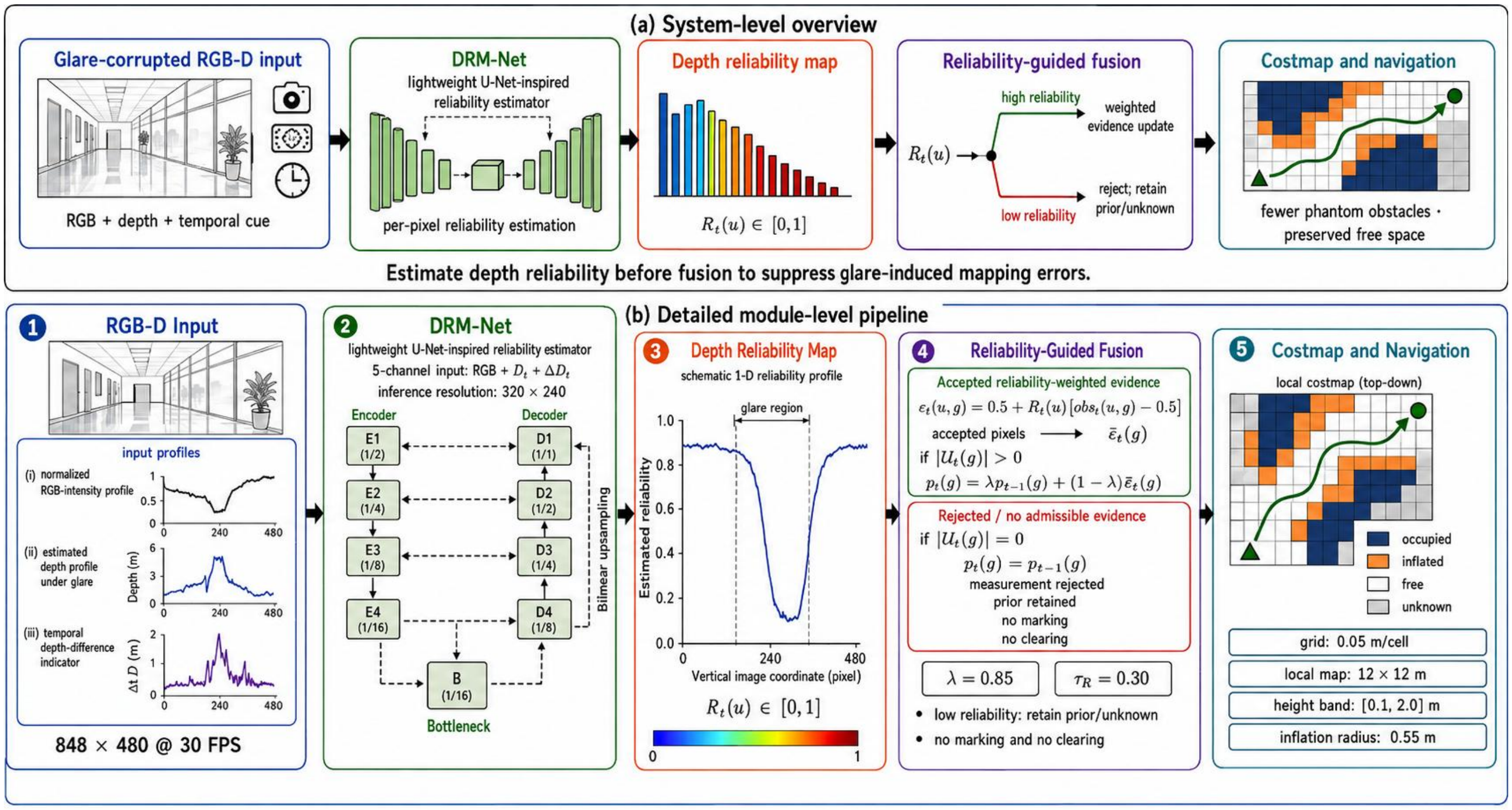

**Fig. 1.** System-level and detailed views of the reliability-guided RGB-D sensor-fusion pipeline. (a) Black-box overview from glare-corrupted RGB-D input through DRM-Net reliability estimation and weighted-and-gated fusion to the navigation costmap. (b) Detailed module-level view showing the five-channel input, reliability profile, fusion gate, and local-costmap integration.

**Contributions—**This work makes five contributions: (1) a glare-aware per-pixel depth-reliability formulation, implemented by a lightweight U-Net-inspired DRM-Net, that predicts active-stereo measurement trustworthiness rather than completed depth; (2) a physically calibrated soft reliability target based on a range-dependent uncertainty function and a five-frame, pose-aligned reference constructed with an independent pose source; (3) a weighted-and-gated occupancy update with explicit unknown-space semantics and a degraded-speed safeguard; (4) a comprehensive comparison against tuned TSDF, Intel RealSense post-processing, stereo-confidence filtering, TDCNet, HDCNet, and learned-target ablations; and (5) a sensor-to-system evaluation reporting exact sample counts, inferential statistics, independent reference maps, collision-related metrics, leave-one-material-out generalization, and embedded runtime.

## II. Related Work

### A. Non-Lambertian Surfaces in RGB-D Sensor Mapping

Specular and transparent materials violate common RGB-D sensing assumptions and can corrupt measurements with missing values, outliers, and view-dependent artifacts. In indoor robotics, reflective ground and glass boundaries degrade mapping and planning by generating phantom structures or by hiding true boundaries. Unlike mapping-level or SLAM-level approaches that primarily repair the resulting map or trajectory estimate, this work targets the local costmap accumulation failure mode by preventing unreliable reflective measurements from being fused into occupancy updates [1]–[4], [9].

### B. RGB-D Depth Completion for Transparent and Reflective Objects

Depth completion is a major direction for transparent and reflective objects. TDCNet uses parallel CNN and Transformer branches for partial-depth and RGB-D features, whereas HDCNet combines CNN, Transformer, and Mamba components for transparent and reflective-object completion [11], [12]. Related benchmarks emphasize dense reconstruction quality [10], [13]–[15]. The evaluation includes TDCNet and HDCNet as quantitative baselines. The claim is not that completion is universally unsuitable; rather, navigation requires density to be evaluated jointly with temporal consistency, false-negative occupancy, and collision-related outcomes.

### C. Reliability, Uncertainty, and Filtering for Sensor Data Fusion

RGB-D calibration and sensor-noise studies show that depth error is heteroscedastic and range dependent [5], [6], [19], [20]. Classical volumetric and probabilistic mapping attenuates uncertain measurements through variance models, robust residuals, or occupancy probabilities [21]–[26]. Recent RGB-D SLAM and stereo-mapping systems further illustrate how depth quality affects dense geometric integration [16], [17]. DRM+RGF follows this reliability-centric view but specializes it for glare-corrupted active-stereo costmaps: a learned per-pixel score is calibrated against multi-view geometry and then used directly in both the weight and admissibility of occupancy updates. The experiments also compare the approach with the D400 SDK filter stack [27].

## D. Mirror and Specular Cues for RGB-D Sensor Reliability

Mirror and specular-region understanding provides an additional cue for identifying unreliable measurements. RGB-D video mirror detection exploits depth and temporal consistency to identify view-dependent failures [28]. More general uncertainty and out-of-distribution cues for monocular depth estimation provide a complementary direction [29]. Although DRM-Net does not explicitly segment mirrors, the reference-depth procedure incorporates multi-view consistency, and the discussion identifies mirror/specular segmentation as a complementary extension.

# III. METHODOLOGY

## A. RGB-D Sensor-Fusion Problem Formulation

Let $D_t(u)$ denote the measured depth at pixel u and time t, $R_t(u) \in [0,1]$ its predicted reliability, and $p_t(g)$ the occupancy probability of grid cell g. Glare can introduce invalid values, holes, and biased disparity spikes that are repeatedly projected into occupied cells. The objective is therefore asymmetric: suppress unsupported occupied evidence without converting missing or rejected measurements into free space. This distinction is central to the safety analysis because an unknown cell is not equivalent to a traversable cell.

For a valid measurement, the 3-D camera-frame point is back-projected using the calibrated intrinsic matrix K and transformed to the map frame with $T_{w\leftarrow c,t}$. Ground-plane quantization produces grid cell g(u). Cells are initialized to a neutral prior and are exported as occupied or free only after the hysteresis thresholds are crossed. If a measurement is rejected by the reliability gate, neither endpoint marking nor ray-based free-space clearing is performed; the affected cell retains its prior state or remains unknown.

## B. DRM-Net Architecture for Depth-Sensor Reliability Output

DRM-Net estimates a per-pixel reliability map $R_{t(u)} \in [0,1]$. Values near 1 indicate geometrically consistent depth measurements, whereas values near 0 indicate likely glare-induced failures. The network operates on downsampled RGB-D input at 320 × 240 in the default five-channel configuration, which includes the temporal depth-difference channel $|D_t - D_{t-1}|$. The RGB-D-only configuration is evaluated only as an ablation. The predicted reliability map is bilinearly upsampled to the native depth resolution before fusion. Table I summarizes the resulting DRM-Net architecture and theoretical computational complexity.

DRM-Net follows the encoder-decoder topology popularized by U-Net [33], but it is not claimed as a new generic backbone. It is a compact, task-specific adaptation that replaces standard convolutions with depthwise-pointwise blocks, uses a reduced 16–128 channel schedule, accepts five-channel RGB-D-temporal input, and predicts a single continuous reliability map through a sigmoid head. Accordingly, the name DRM-Net denotes its reliability-estimation role within the proposed system rather than a claim of a new encoder-decoder family.

**TABLE I**
ARCHITECTURE AND COMPUTATIONAL COMPLEXITY OF DRM-NET

| Block | Type | Output channels | Params | Notes |
|---|---|---|---|---|
| Stem | Conv 3 × 3 | 16 | 720 | 5-channel input: RGB + depth + Δdepth |
| Enc1–4 | DW+PW Conv | 32–128 | 22,864 | Strided downsampling |
| Dec0–3 | Up + DW+PW | 96–16 | 38,336 | Skip connections |
| Head | Conv 1 × 1 + Sigmoid | 1 | 16 | DRM reliability score |
| Total | — | — | 61,936 | ≈0.69 GFLOPs per frame |

*Note: GFLOPs are theoretical architecture-level estimates at 320 × 240. Measured embedded runtime is reported separately in the results section.*

## C. Pose-Aligned Reference Depth and RGB-D Sensor Reliability Targets

The reference depth $D_t^*$ is constructed from a temporal buffer of $N_{buf} = 5$ frames. Historical depth maps are back-projected with the camera intrinsics and reprojected into the current view. Training-reference poses are not estimated from the evaluated D435 stream: they are obtained from an independent 2-D LiDAR/AMCL localization stream fused with wheel odometry. Pose-accuracy validation records gave a translational RMSE of 0.038 m (maximum 0.049 m) and a rotational RMSE of 1.2° (maximum 1.8°), remaining within the 0.05 m and 2° collection limits. This separation reduces the risk that glare-corrupted RGB-D odometry contaminates the reliability labels.

Reprojection uses a z-buffer and a line-of-sight consistency check. A historical sample is accepted only when its depth residual is below $\tau_{res}(z) = 3\sigma(z)$; samples behind the nearest reprojected z-buffer surface are marked as occluded, reprojections more than the 3σ tolerance closer than the current measurement in a majority of valid buffer frames are excluded as transient foreground occluders, and samples with inconsistent depth ordering across the buffer are excluded as dynamic. The accepted frame-index set, support set, and support count at pixel u are

$$\mathcal{A}_t(u) = \{k \in \{t-4, \dots, t\}: \text{sample } k \text{ passes all filtering tests at } u\} \tag{1}$$

$$S_t(u) = \{D_{k\to t}^{reproj}(\mathrm{u}) \mid \mathrm{k} \in \mathcal{A}_t(u)\}. \tag{2}$$

$$N_s(u) = |\mathcal{A}_t(u)| = |S_t(u)| \tag{3}$$

The reference depth is the median over accepted reprojections,

$$D_t^*(u) = \underset{k\in\mathcal{A}_t(u)}{\mathrm{median}} D_{k\to t}^{\mathrm{reproj}}(u), \quad N_s(u) \geq 3 \tag{4}$$

The five-frame window balances viewpoint diversity against dynamic-object ghosting; the reference is undefined where this support criterion is not met.

The range-dependent uncertainty is explicitly modeled as $\sigma(z) = c_0 + c_1 z + c_2 z^2$, with z and σ in meters, $c_0$=0.001063, $c_1$=0.0007278, and $c_2$=0.003949. The quadratic term is consistent with disparity-error propagation in active stereo [6], [20]. DRM-Net is trained with the soft target in (6), which decays with the measured-to-reference residual normalized by $\sigma(D_t^*)$, and with the loss in (7). This prevents naturally larger far-range noise from being labeled as glare while retaining strong penalties for large specular spikes. The complete target-generation procedure is summarized in Fig. 2.

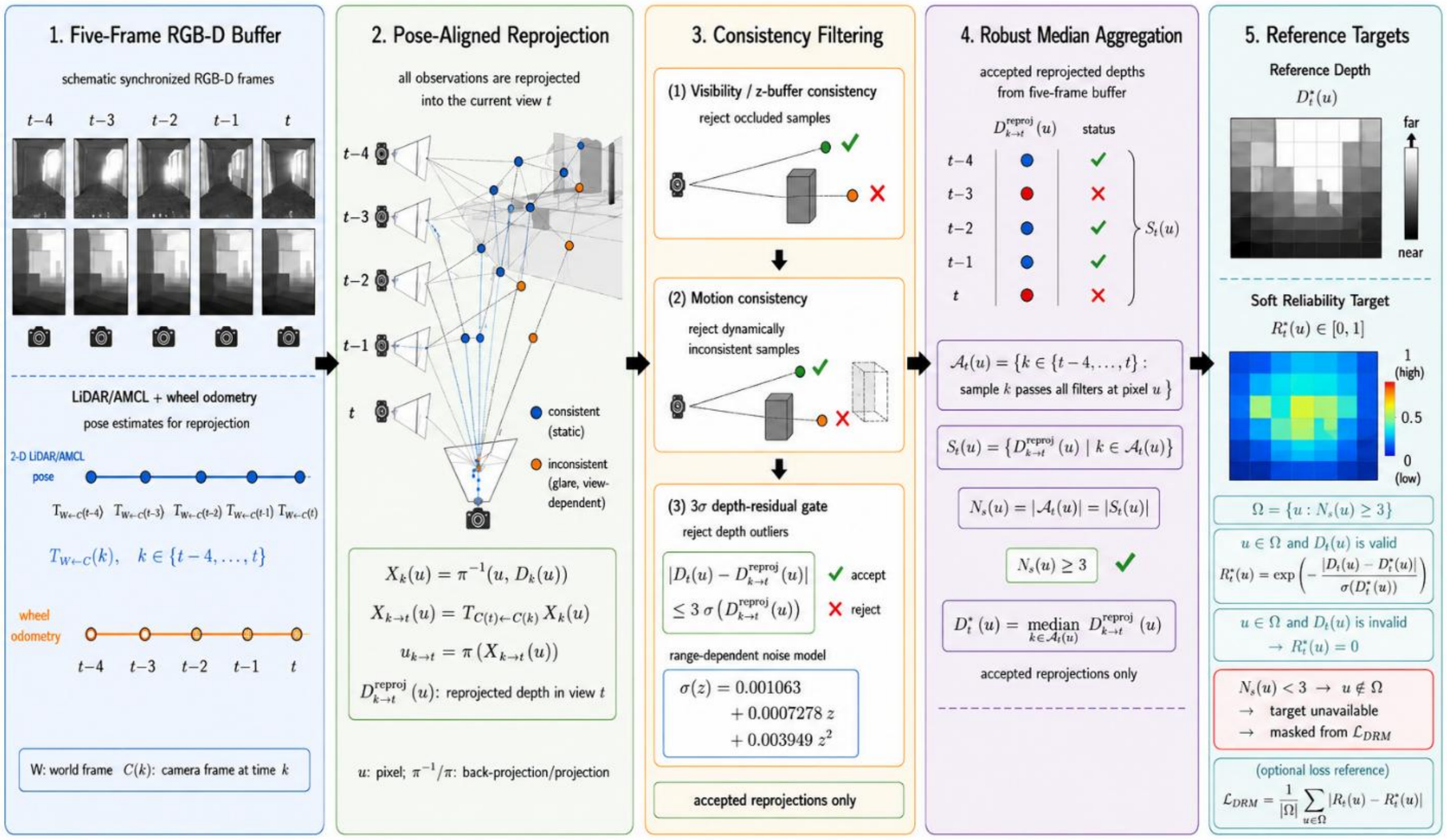

**Fig. 2.** Pose-aligned reliability-target generation for DRM-Net training. Five RGB-D frames are aligned using independent LiDAR/AMCL poses and wheel odometry. After reprojection, visibility, motion-consistency, and 3σ residual filtering, at least three consistent observations are median-aggregated to obtain the reference depth $D_t^*(u)$ and soft reliability target $R_t^*(u)$.

The training-support domain is defined below; $\mathcal{V}_t$ denotes the set of pixels with a valid measured depth $D_t(u)$, i.e., finite and within the configured sensor range.

$$\Omega = \{u\colon N_s(u) \geq 3\} \tag{5}$$

$$R_t^*(u) = \begin{cases} \exp\left(-\dfrac{|D_t(u) - D_t^*(u)|}{\sigma(D_t^*(u))}\right) & u \in \Omega \cap \mathcal{V}_t, \\ 0, & u \in \Omega \setminus \mathcal{V}_t, \\ \text{undefined}, & u \notin \Omega. \end{cases} \tag{6}$$

$$L_{DRM} = \frac{1}{|\,\Omega\,|} \sum_{u\in\Omega} |\, R_t(\mathrm{u}) - R_t^*(\mathrm{u}) \,| \tag{7}$$

*Note: The binary target is used only for ablation and metric thresholding. The support domain is $\Omega = \{u\colon N_s(u) \geq 3\}$. For u∈Ω, an invalid measured depth receives $R_t^*(u) = 0$. Pixels with $N_s(u) < 3$ lie outside Ω, have no reliability target, and are masked from $L_{\text{DRM}}$.*

## D. Reliability-Guided RGB-D Sensor Fusion

RGF combines continuous reliability weighting with a minimum reliability gate. For each grid cell g, only valid pixels satisfying $R_t(u) \geq \tau_R$ are included in $\mathcal{U}_t(g)$. Their reliability-weighted endpoint and ray evidence is averaged to form $\bar{e}_t(g)$. If $\mathcal{U}_t(g)$ is empty, the prior occupancy state is retained and no marking or clearing is performed. The cell-level accepted set and aggregated evidence are

$$\mathcal{U}_t(g) = \{u \in \Omega_t^{\text{valid}}\colon g(u) = g,\ R_t(u) \geq \tau_R\} \tag{8}$$

$$\begin{aligned} e_t(u,g) &= 0.5 + R_t(u)(\text{obs}_t(u,g) - 0.5), \\ \bar{e}_t(g) &= \frac{1}{|\,\mathcal{U}_t(g)\,|} \sum_{u\in\mathcal{U}_t(g)} e_t\,(u,g). \end{aligned} \tag{9}$$

$$p_t(g) = \lambda p_{t-1}(g) + (1-\lambda)\bar{e}_t(g), \quad |\mathcal{U}_t(g)| > 0. \tag{10}$$

$$p_t(g) = p_{t-1}(g), \quad |\mathcal{U}_t(g)| = 0. \tag{11}$$

*Note: λ is the temporal forgetting factor, $\tau_R$ is the minimum reliability threshold, and $\text{obs}_t(u,g) \in [0,1]$ is the endpoint or ray occupancy evidence contributed by pixel u to cell g. Unknown cells are retained as unknown and are configured as non-traversable in Navigation2.*

Unless otherwise stated, λ=0.85 and $\tau_R$=0.3. Both values are fixed after validation and are varied only in the parameter-sensitivity experiment.

The online valid-depth domain is

$$\Omega_t^{\text{valid}} = \{u\colon\ D_t(u) \in [0.17, 10.0]\ \text{m}\} \tag{12}$$

This domain is distinct from the training-support set Ω in (5).

The complete online reliability-guided costmap construction procedure is summarized in Algorithm 1.

**Algorithm 1.** DRM+RGF for Online Glare-Resilient Costmap Construction

```
Inputs: I_t, D_t, |D_t − D_{t−1}|, T_{w←c,t}, p_{t−1}(g)
Outputs: p_t(g) and 2-D costmap
1: R_t ← DRM-Net(I_t, D_t, |D_t − D_{t−1}|)
2: p_t(g) ← p_{t−1}(g) for all grid cells
3: Initialize 𝒰_t(g) ← ∅ for all grid cells
4: for each u ∈ Ω_t^valid do
5:    Project D_t(u) to grid cell g(u)
6:    if R_t(u) ≥ τ_R then
7:       Append pixel u to 𝒰 _t(g(u))
8:    end if
9: end for
10: for each grid cell g do
11:    if |𝒰 _t(g)| > 0 then
12:       Compute ē_t(g) from pixels in 𝒰_t(g) using the cell-level aggregation equation
13:       p_t(g) ← λp_{t−1}(g) + (1−λ)ē_t(g)
14:    end if
15: end for
16: Apply T_on/T_off hysteresis to export costmap
17: return p_t(g), costmap
```

Line 16 applies hysteresis binarization. Rejected measurements do not mark or clear cells, so low reliability cannot be interpreted as free space.

### *E.* Unknown-Space Semantics and Degraded Navigation Mode

The planner treats unknown space conservatively. When the fraction of pixels with $R_t(u) < \tau_R$ exceeds $\theta_{deg} = 0.40$, the controller enters a degraded mode and caps linear speed at $v_{safe} = 0.15$ m/s instead of the nominal $v_{\max} = 0.50$ m/s. The limit is released after the unreliable-pixel ratio falls below $\theta_{deg}$. This policy creates time for small viewpoint changes and repeated observation without asserting free space from missing depth.

## IV. Experimental Setup

### *A. RGB-D Sensor Platform and Mapping Configuration*

Experiments were conducted on a mobile robot equipped with an Intel RealSense D435, a Jetson Orin Nano, wheel odometry, and an independent 2-D LiDAR used for reference localization and mapping. The D435 supplies active-stereo RGB-D at 848×480 and 30 FPS. nvblox integrates depth into a TSDF [26], and the resulting local 12×12 m costmap is consumed by Navigation2 using the fixed costmap and inflation configuration [30], [31]. Every baseline uses the same voxel size, projection band, inflation radius, marking/clearing range, planner configuration, start–goal set, and stopping criteria; only the depth preprocessing or fusion rule changes. Table II summarizes the shared sensing, mapping, costmap, and navigation configuration used across all methods.

**TABLE II**
UNIFIED SENSING, MAPPING, AND COSTMAP CONFIGURATION

| Category | Parameter | Value / setting |
|---|---|---|
| Sensing | RGB-D resolution / rate | 848×480 @ 30 FPS |
| Sensing | Depth range | 0.17–10.0 m |
| Mapping | Voxel / TSDF truncation | 0.05 m / 0.15 m |
| Costmap | Grid / local size | 0.05 m / 12×12 m |
| Costmap | Projection height band | [0.1, 2.0] m |
| Costmap | Inflation radius | 0.55 m |
| Costmap | Mark / clear range | 5.0 m / 5.0 m |
| Costmap | Hysteresis thresholds | $T_{on}$=0.7, $T_{off}$=0.5 |
| Fusion | Default λ / $\tau_R$ | 0.85 / 0.30 |
| Planner | Nominal / degraded speed | 0.50 / 0.15 m/s |
| Planner | Unknown-space policy | Non-traversable |
| Safety | Degraded-mode trigger | Low-reliability ratio >0.40 |

| Category | Parameter | Value / setting |
| --- | --- | --- |
| Trials | Nominal start–goal distance | 10 m |
| Trials | Valid trials by glare level | L0 100/102; L1 92/111; L2 98/105 (Baseline/DRM+RGF) |
| Outcome | Success / failure | Goal reached within the timeout / collision, unrecoverable stop, or timeout |

*Note: Exact Navigation2 and nvblox YAML parameters are held fixed across methods. Exact valid-trial denominators are reported for L0–L2; the prespecified inferential analysis uses the L2 protocol, whereas L0 and L1 remain descriptive condition checks. Safety interventions are recorded separately and do not by themselves define navigation failure.*

### B. Glare Scenarios and Sensor Protocol

Glare severity is evaluated at L0 (no glare), L1 (mild glare), and L2 (severe glare). The test set covers polished floor corridors, glass walls/automatic doors, and glossy tiles under natural window light. These conditions induce distinct active-stereo failures, including multipath reflection, IR-pattern penetration or redirection, and sunlight-induced IR saturation. Start–goal orientations are varied so that the sensor approaches reflective surfaces from multiple angles.

### C. Fusion Baselines and Evaluation Metrics

The comparison includes tuned nvblox TSDF, validity/range gating, spatial median and temporal rejection, the Intel RealSense SDK spatial-temporal-hole-filling stack, high-threshold stereo-confidence filtering, TDCNet, HDCNet, and DRM+RGF. In addition, a binary-target DRM variant serves as a learned per-pixel depth-confidence baseline. For aligned prediction and independent-reference grids, let $O_{pred}$ and $F_{pred}$ denote predicted occupied and free cells, and $O_{ref}$ and $F_{ref}$ denote reference occupied and free cells. Define the known-reference domain $K_{ref} = O_{ref} \cup F_{ref}$ . The occupancy-confusion records use TPO= $|O_{pred} \cap O_{ref}|$ , FPO= $|O_{pred} \cap F_{ref}|$ , TPF= $|F_{pred} \cap F_{ref}|$ , and FNO= $|F_{pred} \cap O_{ref}|$ ; FOR=FPO/(FPO+TPO), FSR=TPF/$|F_{ref}|$, and FNOR=FNO/$|O_{ref}|$. Predicted cells aligned to reference-unknown cells lie outside $K_{ref}$ and are excluded rather than reinterpreted as free. A representative aligned-grid audit on the 12×12 m, 0.05 m/cell evaluation grid (57,600 cells) reproduces the reported FOR, FSR, and FNOR of trials B_01 and D_01 from raw confusion counts and is included in the released records. Navigation outcomes include safety interventions per 10 m, success, path-length ratio, time-to-goal, collisions, and near-collisions (< 0.15 m). Safety interventions are protocol-defined protective controller actions: Baseline events are emergency-stop/recovery events, whereas DRM+RGF events are entries into the 0.15 m/s degraded mode and are not zero-velocity stops.

### D. RGB-D Sensor Dataset and Implementation Details

The in-house RGB-D dataset comprises 52 sequences from three indoor environments and approximately 21,600 reliability-labeled frames, split by sequence into 36/8/8 training, validation, and test sets to prevent trajectory overlap. Approximately 100 trials were scheduled per method and glare condition. Runs with invalid logs due to mobile-base initialization failure or telemetry loss were excluded independently of navigation outcome. Successes/retained valid trials (Baseline vs. DRM+RGF) were 100/100 vs. 102/102 at L0, 75/92 (81.5%) vs. 109/111 (98.2%) at L1, and 28/98 (28.6%) vs. 96/105 (91.4%) at L2. Counts may exceed 100 because this was an approximate target rather than a fixed quota. All 203 retained L2 trials followed the same preplanned protocol; none was added as a separate edge case. Each trial records success, collisions, near-collision incidence and count, minimum clearance, safety interventions, travel distance, FOR, FSR, FNOR, and navigation outcomes. Full-sample inference uses Baseline N=98 and DRM+RGF N=105, while path-length ratio and time-to-goal use the 28 start–goal pairs completed by both methods; zero-coded failure placeholders are excluded. Exact integer denominators are reported throughout.

DRM-Net is trained for 60 epochs with Adam ($\beta_1 = 0.9$, $\beta_2 = 0.999$, weight decay $1 \times 10^{-4}$), batch size 8, and cosine learning-rate decay from $1 \times 10^{-3}$ to $1 \times 10^{-5}$. The reliability gate is selected by validation F1 and fixed at $\tau_R$=0.3. Horizontal flipping is the only augmentation, and all metrics are computed from frozen test outputs.

### E. Independent Ground Truth and Statistical Protocol

FOR, FSR, and FNOR are evaluated against a reference map independent of the D435 stream. A 2-D LiDAR SLAM map is registered to manually surveyed glass boundaries, metallic columns, and reflective wall edges using 15 control points and a unit-scale rigid 2-D Umeyama alignment [32]. With $x_{survey}$ and $x_{lidar}$ denoting survey and LiDAR-map coordinates, the 15 recorded control-point pairs yield $x_{lidar} = Rx_{survey} + t$ with R=[[0.999999714, 0.000756729], [-0.000756729, 0.999999714]] and t=[-0.001935, 0.005266]$^{T}$ m, corresponding to an in-plane rotation of -0.0434°. Pointwise residuals give an RMSE of 0.0139 m and a maximum residual of 0.0197 m. The fixed geometry is then projected into the evaluation costmap. Full-sample continuous outcomes are summarized as mean±standard deviation with 95% confidence intervals and compared using two-sided Welch's t-tests when within-group variance is nonzero. FNOR is reported descriptively because the trial table contains a constant FNOR value within each method. Path-length ratio and time-to-goal use paired t-tests on the 28 matched completed pairs; binary success and safety incidences use two-sided Fisher exact tests. FOR, FSR, safety interventions/10 m, and global success are treated as the primary L2 system-level outcomes in this paper; remaining tests are secondary and exploratory, and their p values are unadjusted. The independent reference-costmap construction is summarized in Fig. 3.

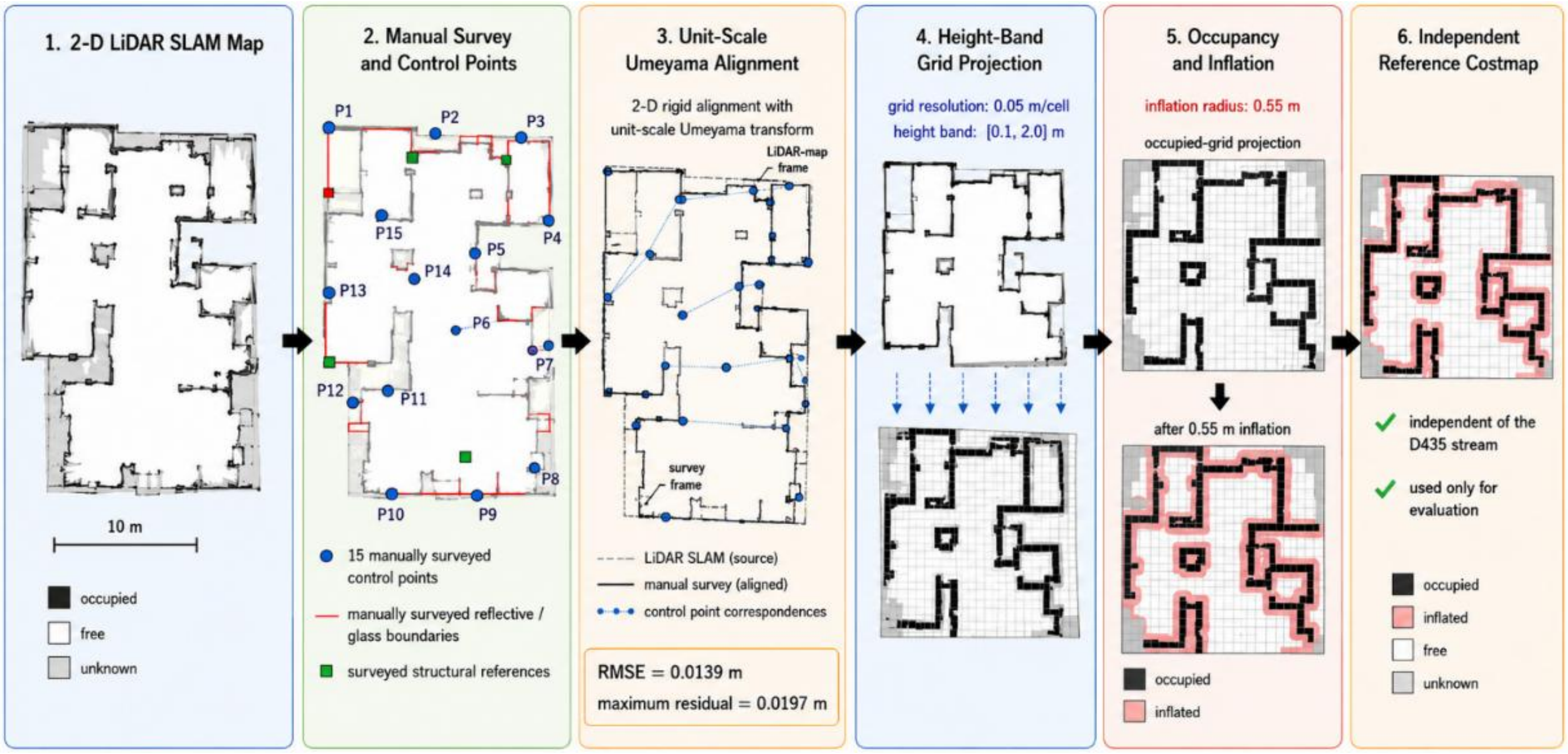

**Fig. 3.** Independent reference-costmap construction for evaluation. A 2-D LiDAR SLAM map is aligned to manually surveyed reflective boundaries through a unit-scale Umeyama alignment on the 15 recorded control-point pairs (RMSE 0.0139 m, maximum 0.0197 m), projected onto a 0.05 m/cell grid within the [0.1, 2.0] m height band, and inflated by 0.55 m. The panels illustrate processing stages; the resulting reference costmap is independent of the D435 stream and is used only for evaluation.

## V. Results and Analysis

### A. RGB-D Depth-Sensor Corruption Suppression

Table III reports depth-stream corruption before and after DRM filtering. DRM suppresses unreliable glare-corrupted measurements before costmap accumulation, reducing erroneous-depth rate (EDR, %), spike acceptance, and retained-depth RMSE within the admissible evidence stream, especially under severe glare.

**TABLE III**
RESIDUAL CORRUPTION IN ADMISSIBLE DEPTH EVIDENCE

| Glare level | Evidence stream | EDR (%) ↓ | Spike acceptance rate (%) ↓ | RMSE of retained valid depth (m) ↓ |
|---|---|---|---|---|
| L0 (no glare) | Raw D435 depth stream | 1.2 | 0.5 | 0.022 |
| L0 (no glare) | DRM-filtered depth stream | 1.1 | 0.4 | 0.021 |
| L1 (mild) | Raw D435 depth stream | 18.5 | 9.4 | 0.185 |
| L1 (mild) | DRM-filtered depth stream | 2.8 | 1.2 | 0.045 |
| L2 (severe) | Raw D435 depth stream | 45.2 | 27.6 | 0.840 |
| L2 (severe) | DRM-filtered depth stream | 5.4 | 3.1 | 0.082 |

*Note: The navigation-trial counts in Table II are not the denominators of the depth-stream statistics. Both raw and DRM-filtered rates are computed over finite, in-range samples after validity screening; DRM-filtered rates additionally apply the reliability gate. EDR denotes the fraction of admitted samples whose depth disagrees with the independent reference according to the evaluation criterion. Exact evaluated frame and pixel denominators are unavailable in the current records and are therefore not inferred from navigation-trial counts. These rates do not measure full-image depth completion.*

### B. Reliability-Target and Reference-Construction Ablation

Reliability-map quality is evaluated using AUPRC and F1 because glare-corrupted pixels are sparse and class imbalance makes raw accuracy misleading. Table IV further evaluates whether the final reliability estimator depends on the soft target, the temporal depth-difference input, and the pose-aligned multi-view reference-depth construction. Compared with binary supervision, the soft target avoids unstable hard labels near glare boundaries. The temporal term improves detection of view-dependent depth failures, while multi-view reference construction reduces the chance that persistent glare artifacts are treated as valid supervision. The default configuration therefore uses soft reliability supervision, RGB-D plus temporal depth difference, and pose-aligned multi-view reference construction in all main experiments. For context across glare levels, Fig. 4 summarizes the recorded Baseline TSDF and DRM+RGF system-level outcomes at L0–L2. For qualitative interpretation only, Fig. 5 provides a schematic, non-recorded illustration of the costmap and path behavior associated with the two methods.

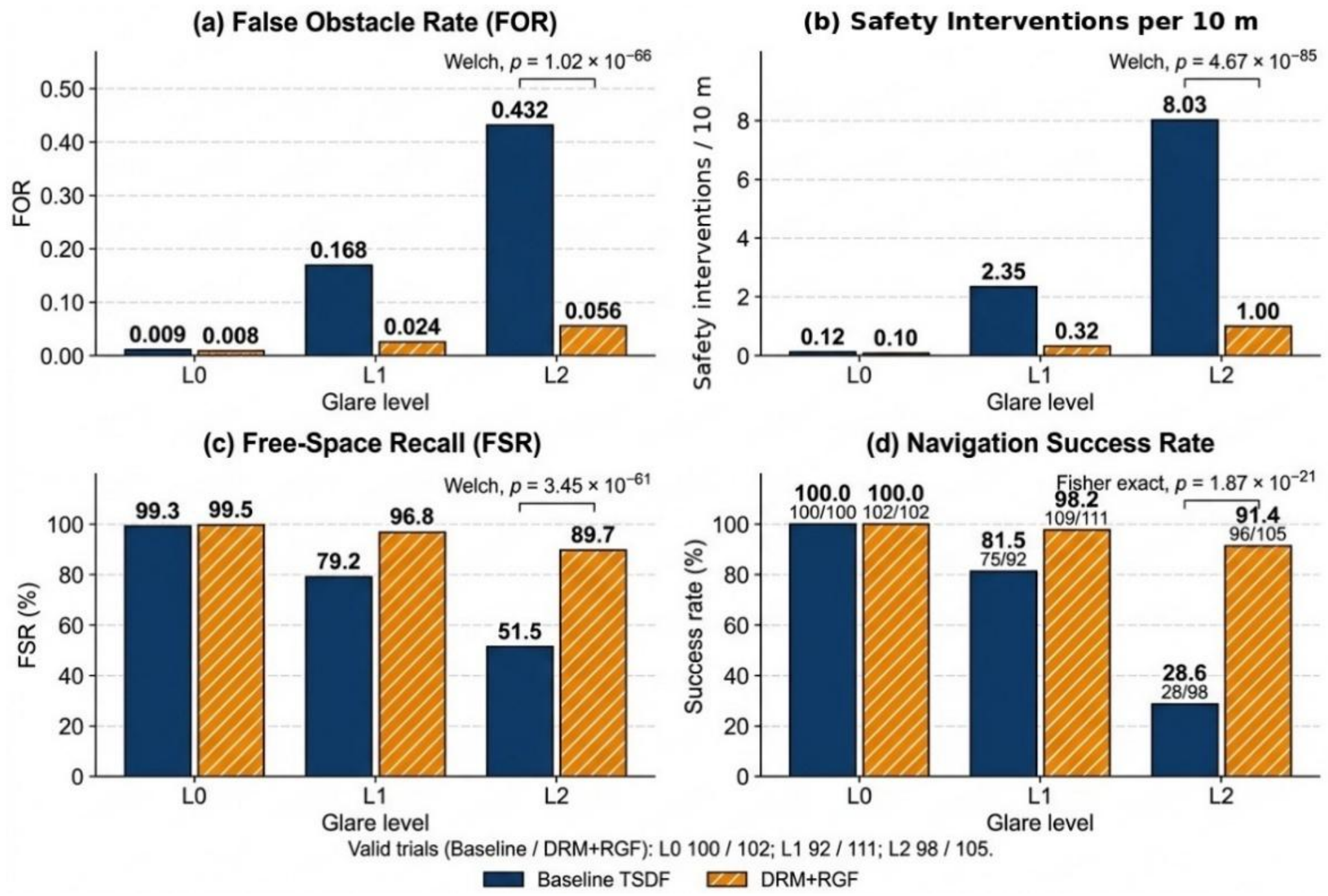

**Fig. 4.** Recorded L0–L2 results for Baseline TSDF and DRM+RGF. Valid trial counts by glare level are given in Table II. Bars show FOR, safety interventions, FSR, and global success; the L2 statistical annotations correspond to the tests reported in Table IX. (a) FOR. (b) Safety interventions per 10 m. (c) FSR. (d) Navigation success rate.

**TABLE IV**
RELIABILITY-TARGET AND REFERENCE-CONSTRUCTION ABLATION

| Target / input / reference setting | AUPRC ↑ | F1 ↑ | L2 FOR ↓ | L2 FSR ↑ | Safety interventions/10 m ↓ | Success (%) ↑ | Observation |
|---|---|---|---|---|---|---|---|
| Binary target + Δdepth + multi-view ref. | 0.81 | 0.78 | 0.089 | 0.843 | 1.38 | 83.0 (78/94) | Unstable near glare boundaries |
| Soft target + RGB-D only + multi-view ref. | 0.84 | 0.82 | 0.074 | 0.862 | 1.15 | 86.8 (79/91) | Temporal cue removed |
| Soft target + Δdepth + temporal-filter ref. only | 0.73 | 0.70 | 0.121 | 0.806 | 2.08 | 72.5 (74/102) | Circular supervision risk |
| Soft target + Δdepth + multi-view ref. (default) | 0.89 | 0.87 | 0.056 | 0.897 | 1.00 | 91.4 (96/105) | Default; best overall |

*Note: AUPRC and F1 are computed on the held-out reliability test set; F1 uses the validation-selected threshold. Exact frame- and pixel-level denominators are unavailable in the retained records.*

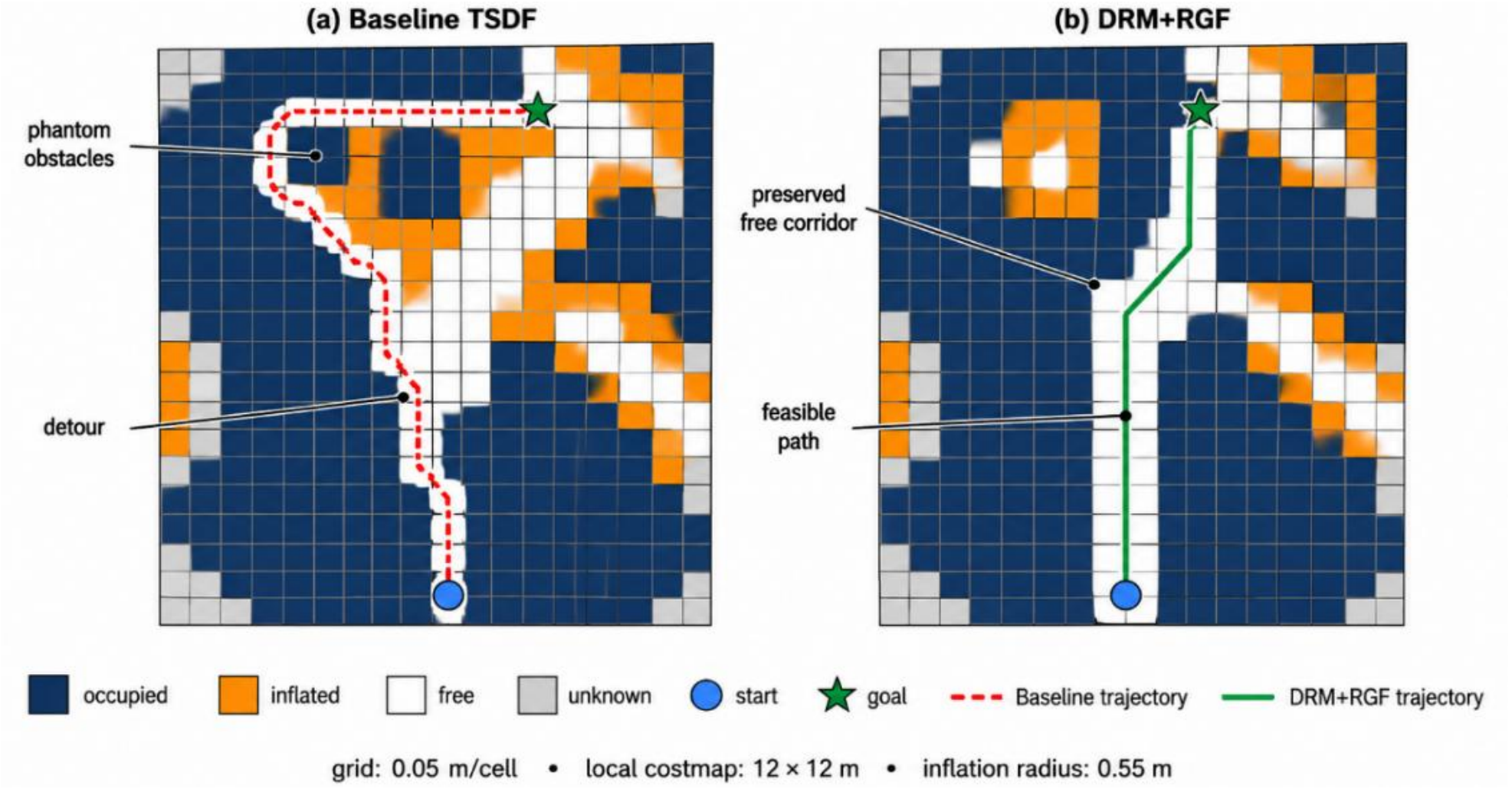

**Fig. 5.** Schematic comparison of navigation behavior under severe-glare-inspired costmap conditions. (a) Baseline TSDF exhibits phantom occupied regions and follows a longer detour. (b) DRM+RGF preserves the free corridor and supports a shorter feasible path. The costmaps and trajectories are schematic visualizations derived from the recorded costmap grid structure and are not recorded planner outputs.

## *C. Costmap Correctness and Navigation Performance*

Table V presents the L2 comparison across filtering, confidence-based rejection, depth completion, and reliability-guided fusion methods.

**TABLE V**
BASELINE AND SYSTEM-LEVEL COMPARISON UNDER L2 GLARE

| Method | FOR ↓ | FSR ↑ | FNOR ↓ | Safety interventions/ 10 m ↓ | Success ↑ | Latency |
|---|---|---|---|---|---|---|
| nvblox TSDF | 0.432 | 0.515 | 0.012 | 8.03 | 28.6% (28/98) | 10.0 ms |
| RealSense SDK stack | 0.385 | 0.590 | 0.015 | 5.92 | 34.5% (29/84) | 14.5 ms |
| High stereo confidence | 0.085 | 0.410 | 0.008 | 2.15 | 55.1% (49/89) | 10.5 ms |
| TDCNet | 0.115 | 0.825 | 0.084 | 1.85 | 72.0% (67/93) | 45.2 ms |
| HDCNet | 0.092 | 0.860 | 0.065 | 1.42 | 81.5% (66/81) | 85.0 ms |
| DRM+RGF | 0.056 | 0.897 | 0.018 | 1.00 | 91.4% (96/105) | 16.5 ms |

Across the 203 retained L2 main-comparison trials, the Baseline TSDF and DRM+RGF rows in Table V reproduce the rounded FOR and FSR values and yield 8.03±0.99 versus 1.00±0.00 safety interventions per 10 m (Δ=-7.03, 95% CI [-7.23, -6.83], Welch t=-70.36, df=97, $p=4.67\times10^{-85}$). Stereo-confidence filtering lowers FNOR but sacrifices free-space evidence, whereas completion improves FSR at higher FNOR and latency. DRM+RGF achieves the lowest FOR and intervention rate and the highest success while keeping FNOR near the TSDF baseline, supporting a balanced tradeoff rather than rejecting depth completion outright.

### *D. Sensor-Fusion Variant Ablation*

Table VI separates the effects of weighting and gating. Weighted-only fusion attenuates errors but permits persistent low-weight accumulation; gate-only fusion reduces false obstacles but retains less edge geometry. The combined rule provides the highest success and FSR while keeping FOR low.

### *E. Parameter Sensitivity Analysis*

Table VII evaluates λ and $\tau_R$. The default λ=0.85 and $\tau_R = 0.3$ lies in the stable region. Smaller λ reacts quickly but increases map flicker; larger λ retains corrupted evidence. A lower gate admits more unreliable measurements, whereas a higher gate blocks useful edge geometry.

### *F. Cross-Condition Evaluation*

Table VIII reports performance across three overlapping glare-dominant scene categories, so a single 10 m trial may contribute to multiple rows. In leave-one-material-out testing, withholding the glass-wall/automatic-door scene during training reduced AUPRC from 0.89 to 0.85 and increased L2 FOR from 0.056 to 0.072, still well below the TSDF FOR of 0.432. This indicates transfer across reflective materials, but not cross-sensor generality.

**TABLE VI**
FUSION-VARIANT ABLATION UNDER L2 GLARE

| Fusion variant | FOR ↓ | FSR ↑ | Safety interventions/10 m ↓ | Success (%) ↑ | Main observation |
|---|---|---|---|---|---|
| Baseline TSDF | 0.432 | 0.515 | 8.03 | 28.6 (28/98) | Phantom wall accumulation |
| Weighted-only | 0.185 | 0.724 | 3.15 | 65.2 (60/92) | Artifacts persist |
| Gate-only | 0.048 | 0.785 | 1.10 | 85.5 (94/110) | Overly conservative |
| Weighted+Gate | 0.056 | 0.897 | 1.00 | 91.4 (96/105) | Best balance |

**TABLE VII**
PARAMETER SENSITIVITY UNDER L2 GLARE

| λ | $\tau_R$ | FOR ↓ | FSR ↑ | Safety interventions/10 m ↓ | Success (%) ↑ | Behavior |
|---|---|---|---|---|---|---|
| 0.75 | 0.3 | 0.082 | 0.915 | 1.85 | 75.5 (74/98) | Map flicker |
| 0.95 | 0.3 | 0.145 | 0.810 | 2.60 | 62.0 (67/108) | Over-persistent occupancy |
| 0.85 | 0.2 | 0.110 | 0.920 | 1.95 | 80.0 (84/105) | Gate too permissive |
| 0.85 | 0.4 | 0.035 | 0.765 | 1.15 | 82.5 (80/97) | Gate too conservative |
| 0.85 | 0.3 | 0.056 | 0.897 | 1.00 | 91.4 (96/105) | Stable |

### *G. Paired Completed-Trial Navigation Comparison*

Because path and time metrics can be biased when one method fails many trials, Table IX analyzes the 28 start–goal pairs completed by both methods and reports global success separately on the full samples. Pairs were matched by shared start–goal identifiers. PLR decreased from 1.920 ± 0.458 to 1.061 ± 0.040, with a paired DRM+RGF-minus-Baseline difference of -0.859 ± 0.464 (t=-9.81, df=27, 95% CI [−1.039, −0.680], $p=2.16\times10^{-10}$). Time-to-goal decreased from 85.30±18.53 s to 48.50 ± 3.16 s, with a paired difference of −36.80 ± 18.50 s (t=-10.52, df=27, 95% CI [−43.97, −29.63] s, $p=4.69\times10^{-11}$). These paired results characterize efficiency only among jointly completed routes and do not replace the full-sample success comparison.

**TABLE VIII**
CROSS-CONDITION EVALUATION ACROSS GLARE SCENARIOS

| Scene | Optical condition | Method | FOR ↓ | FSR ↑ | Safety interventions/10 m ↓ | Success (%) ↑ |
|---|---|---|---|---|---|---|
| Reflective floor corridor | Polished epoxy, overhead LED | Baseline TSDF | 0.380 | 0.610 | 5.50 | 35.0 (14/40) |

| Scene | Optical condition | Method | FOR ↓ | FSR ↑ | Safety interventions/10 m ↓ | Success (%) ↑ |
|---|---|---|---|---|---|---|
| Reflective floor corridor | Polished epoxy, overhead LED | DRM+RGF | 0.045 | 0.920 | 0.70 | 94.5 (69/73) |
| Glass wall / automatic door | Transparent glass, side diffuse light | Baseline TSDF | 0.295 | 0.585 | 4.80 | 42.5 (17/40) |
| Glass wall / automatic door | Transparent glass, side diffuse light | DRM+RGF | 0.068 | 0.865 | 1.10 | 88.0 (44/50) |
| Glossy tile with window light | IR saturation from natural light | Baseline TSDF | 0.492 | 0.415 | 8.20 | 15.5 (9/58) |
| Glossy tile with window light | IR saturation from natural light | DRM+RGF | 0.075 | 0.812 | 1.45 | 85.5 (47/55) |

*Note: Scene categories are assigned along each trial route, and scene regions can physically overlap within a single traverse, so one trial can contribute to multiple scene rows. Per-scene denominators therefore sum to more than the pooled L2 totals and do not partition the sample; per-scene values are descriptive and are not used for between-scene inference.*

**TABLE IX**
STATISTICAL COMPARISON UNDER L2 GLARE

| Metric | Baseline | DRM+RGF | Inference / 95% CI |
|---|---|---|---|
| FOR | 0.432±0.085 | 0.056±0.012 | Δ=-0.376 [-0.393, -0.359]; t=-43.16, df=100.59, $p=1.02\times10^{-66}$ |
| FSR | 0.515±0.113 | 0.897±0.045 | Δ=0.382 [0.358, 0.406]; t=31.33, df=125.54, $p=3.45\times10^{-61}$ |
| Safety interventions/10 m | 8.03±0.99 | 1.00±0.00 | Δ=-7.03 [-7.23, -6.83]; t=-70.36, df=97, $p=4.67\times10^{-85}$ |
| Paired PLR (n=28) | 1.920±0.458 | 1.061±0.040 | Δ=-0.859±0.464 [-1.039, -0.680]; t=-9.81, $p=2.16\times10^{-10}$ |
| Time-to-goal (n=28) | 85.30±18.53 s | 48.50±3.16 s | Δ=-36.80±18.50 s [-43.97, -29.63]; t=-10.52, $p=4.69\times10^{-11}$ |
| Global success | 28/98 (28.6%) | 96/105 (91.4%) | Baseline Wilson [20.6%, 38.2%]; DRM+RGF Wilson [84.5%, 95.4%]; Fisher $p=1.87\times10^{-21}$ |

*Note: FOR, FSR, and safety interventions/10 m are computed from all 203 retained L2 trials and use Welch's two-sample t-test; reported differences are DRM+RGF minus Baseline. FNOR is descriptive because it has zero within-method variance in the recorded trials. Paired PLR and time-to-goal use the 28 jointly completed pairs; their differences are likewise DRM+RGF minus Baseline, and failure-row zero placeholders are excluded. Wilson intervals are reported for both global-success proportions, and the two-sided Fisher exact test gives $p=1.87\times10^{-21}$. Full-sample L2 inference uses Baseline N=98 and DRM+RGF N=105. Paired path-length ratio and time-to-goal comparisons use N=28 jointly completed pairs.*

## H. Sensor-Fusion Runtime and Embedded Deployment

DRM-Net was exported through ONNX and executed as a TensorRT FP16 engine on the Jetson Orin Nano in 15 W mode with 320 × 240 input. Timing used ROS 2 message_filters::sync::ApproximateTime synchronization. After excluding the first 50 warm-up frames, the retained timing record contains 1000 frames. Mean DRM inference latency was 6.44 ms. Complete-pipeline latency was 16.50 ± 1.45 ms (mean ± SD), with a median of 16.4 ms, P95 of 18.7 ms, P99 of 21.0 ms, and a maximum of 23.6 ms. Even the observed maximum remained below the 33.3 ms D435 frame period. The logged fusion and costmap-export components averaged 4.62 and 1.98 ms, respectively; the remaining 3.46 ms of mean pipeline overhead is the difference between the total and the three explicitly logged components. Table X summarizes the component and end-to-end measurements.

**TABLE X**
RUNTIME AND COMPUTATIONAL OVERHEAD

| Method | DRM (ms) | Fusion + export (ms) | Other (ms) | Total (ms) | FPS |
|---|---|---|---|---|---|
| Baseline TSDF | — | — | — | 10.0 | 100.0 |
| DRM+RGF | 6.44 | 6.60 | 3.46 | 16.5 | 60.6 |

*Note: GFLOPs in Table I are theoretical; runtime statistics use the 1000 post-warm-up frames described in the text. Baseline component-level timing was not logged separately; only its complete-pipeline total is reported.*

## I. Safety Evaluation and Degraded-Mode Behavior

In the retained high-risk reflective-scene trials (Baseline N=98; DRM+RGF N=105), DRM+RGF produced a small descriptive FNOR increase relative to raw TSDF (0.018 versus 0.012) but substantially improved the observed safety outcomes. Mean minimum clearance increased from 0.193 ± 0.137 m to 0.385 ± 0.069 m (Δ = 0.192 m, 95% CI [0.162, 0.222], Welch t=12.48, df=141.16, $p=1.46\times10^{-24}$). Collisions decreased from 14/98 to 1/105. At least one near-collision occurred in 49/98 Baseline trials and 5/105 DRM+RGF trials; the corresponding event totals were 147 and 5. The supplementary sample of eight timestamped intervention events (seven Baseline emergency-stop/recovery events in B_01 and one DRM+RGF degraded-mode entry in D_01) demonstrates event-to-telemetry alignment; full-sample inference uses the complete 203-trial record. Table XI summarizes the FNOR, collision, and near-collision outcomes.

**TABLE XI**
SAFETY STRESS TEST AND DEGRADED-MODE RESPONSE

| Outcome | Baseline / normal | DRM+RGF / degraded | Valid trials | Statistical result |
|---|---|---|---|---|
| FNOR | 0.012 | 0.018 | 98 / 105 | Δ=0.006 |
| Collisions | 14/98 | 1/105 | 98 / 105 | $p=2.26\times10^{-4}$ |
| Trials with near-collision | 49/98 | 5/105 | 98 / 105 | $p=4.14\times10^{-14}$ |

*Note: A near-collision is a minimum obstacle clearance below 0.15 m; the row counts trials with at least one event. Collision and near-collision incidence use two-sided Fisher exact tests.*

The zero within-method variance of several DRM+RGF fields reflects the controlled L2 stress-test protocol rather than zero uncertainty in general operation: static reflective geometry, frozen DRM-Net weights, and a repeated 10 m approach produce identical FNOR counts, and the deterministic low-reliability trigger (>0.40 of the field of view) yields exactly one 3.5 s degraded-mode speed-reduction entry per traverse. Such quantities are reported descriptively. The 0.50 and 0.15 m/s values in Table II are controller speed caps, whereas the 0.20 m/s mean speed in the trial logs is a protocol-level summary rather than an inferential outcome.

## VI. Discussion

### A. Why Reliability-Guided Sensor Fusion Instead of Full Depth Completion?

The baseline results qualify the paper's central argument. Completion is beneficial when dense geometry is the primary objective, and both TDCNet and HDCNet improve FSR; in the integrated pipeline, however, TDCNet's smoothing can suppress thin obstacle structure and HDCNet's frame-wise completion can flicker across viewpoint changes, consistent with their elevated L2 FNOR values (0.084 and 0.065), longer latency, and repeated local replanning. Conversely, strict stereo-confidence filtering lowers FNOR to 0.008 but leaves too much unknown space (FSR 0.410). DRM+RGF occupies the middle ground: it does not invent geometry, yet it preserves graded evidence when the sensor remains trustworthy, and its small FNOR increase (0.012 to 0.018) indicates that conservative rejection does not eliminate the risk of missed occupancy evidence.

### B. Circular Supervision and Reference-Depth Validity

Circular supervision is addressed at two levels: reference-depth labels use an independent LiDAR/AMCL pose stream (0.038 m translational and 1.2° rotational RMSE) with five-frame reprojection, occlusion handling, and a $3\sigma$ residual test, and downstream FOR/FSR/FNOR metrics use the survey-registered LiDAR map of Section IV-E, which is independent of both the D435 input and the online fusion output. These controls substantially reduce, but do not eliminate, label error at transparent boundaries.

### C. Failure Cases and Degraded Sensing

The method can still under-insert obstacles when most of the field of view is persistently unreliable, including large mirrors, wet floors, and glass façades. The system therefore retains rejected regions as unknown and activates a 0.15 m/s re-observation mode when the low-reliability ratio exceeds 0.40. The safety stress test shows fewer collisions and near-collisions, but these results are platform- and environment-specific and should not be interpreted as formal safety certification.

### D. Extensions to Multi-Sensor Robustness

The leave-one-material-out result supports scene transfer within D435 active-stereo failure modes, but all primary experiments still use one depth sensor and one embedded platform. A time-of-flight sensor would exhibit different multipath and range-noise behavior and would require recalibration of $\sigma(z)$ and potentially different input channels. The weighting-and-gating rule is sensor agnostic, whereas the learned reliability estimator is not assumed to transfer without adaptation. Future work will evaluate cross-sensor calibration and RGB-D/LiDAR/polarization reliability fusion [8].

## VII. Conclusion

This paper presents a reliability-guided RGB-D fusion framework for suppressing glare-corrupted active-stereo measurements before they propagate into TSDF mapping and navigation costmaps. The reliability target is specified through a calibrated quadratic uncertainty model, independent-pose multi-view references, occlusion-aware aggregation, and explicit unknown-space semantics. The evaluation also includes SDK filtering, stereo-confidence filtering, and two depth-completion networks.

Under severe glare, DRM+RGF reduces FOR from 0.432 to 0.056, increases FSR from 0.515 to 0.897, reduces protective safety interventions from 8.03 Baseline emergency-stop/recovery events to 1.00 DRM+RGF degraded-mode entry per 10 m, and raises global success from 28.6% (28/98) to 91.4% (96/105; Fisher $p=1.87\times10^{-21}$), with a small descriptive FNOR increase from 0.012 to 0.018. Among the 28 jointly completed routes, path-length ratio decreases by 0.859 and time-to-goal by 36.8 s on average, collisions decrease from 14/98 to 1/105, and the complete pipeline runs in 16.5 ms (maximum 23.6 ms) on the embedded platform. These findings support reliability-guided fusion as a practical safety-utility tradeoff for D435-based indoor navigation, while cross-sensor validation remains future work.

## Data and Code Availability

The GlaReNav-RGBD reproducibility package is available at https://github.com/seanideslab/GlaReNav-RGBD. It includes the 203 retained L2 trial records, the 28-pair subset, representative occupancy and safety records, 15 registration pairs, 1000 post-warm-up runtime samples, table exports, validation scripts, a repository manifest, and SHA-256 checksums. The complete raw RGB-D archive and ROS bags are not included.